\documentclass[11pt]{article}

\usepackage{arxiv}

\usepackage{fancyhdr,hyperref, setspace,microtype,csquotes,placeins,url,soul,booktabs}
\usepackage{amsmath,amsfonts,amssymb,amsthm}
\usepackage{algorithm, algorithmic}
\usepackage{subfigure,graphicx,color}

\usepackage[small]{caption}
\usepackage[normalem]{ulem}
\usepackage[english]{babel}
\usepackage{wrapfig}
\title{Policy-focused Agent-based Modeling\\ using RL Behavioral Models}
\author{
    Osonde A. Osoba\footnotemark\\
    \And
    Raffaele Vardavas \\
    \And
    Justin Grana \\
    \And
    Rushil Zutshi \\
    \And
    Amber Jaycocks \\
}

\begin{document}
\maketitle
\footnotetext{Corresponding Author Email: oosoba@rand.org}

\begin{abstract}
	Agent-based Models (ABMs) are valuable tools for policy analysis. ABMs help analysts explore the emergent consequences of policy interventions in multi-agent decision-making settings. But the validity of inferences drawn from ABM explorations depends on the quality of the ABM agents' behavioral models. Standard specifications of agent behavioral models rely either on heuristic decision-making rules or on regressions trained on past data. Both prior specification modes have limitations. This paper examines the value of reinforcement learning (RL) models as adaptive, high-performing, and behaviorally-valid models of agent decision-making in ABMs. We test the hypothesis that RL agents are effective as utility-maximizing agents in policy ABMs. We also address the problem of adapting RL algorithms to handle multi-agency in games by adapting and extending methods from recent literature. We evaluate the performance of such RL-based ABM agents via experiments on two policy-relevant ABMs: a minority game ABM and an ABM of Influenza Transmission. We run some analytic experiments on our AI-equipped ABMs e.g. explorations of the effects of behavioral heterogeneity in a population and the emergence of synchronization in a population. The experiments show that RL behavioral models are effective at producing reward-seeking or reward-maximizing behaviors in ABM agents. Furthermore, RL behavioral models can learn to outperform the default adaptive behavioral models in the two ABMs examined.
	These results suggest that the RL formalism can be an efficient default abstraction for behavioral models in ABMs. The core of the argument is that the RL formalism cleanly and efficiently represents reward-seeking behavior in intelligent agents. Furthermore, the RL formalism allows modelers to specify behavioral models in a way that is agnostic to the agent’s detailed internal structure of decision-making, which is often unidentifiable.
\end{abstract}

\section{Introduction: Why RL for ABMs?}
\label{intro}
Agents based models (ABMs) are useful for analyzing the dynamics of complex social systems~\cite{blume2015agent,epstein2006generative}.
In its most abstracted form, an ABM consists of virtual agents interacting with each other in a virtual environment.
ABMs are natural tools for policy analysis.
A good ABM can enable policy analysts and decision-makers to explore the potential macro-effects of diverse policy interventions on a population of real-world agents.

Modelers will typically equip each agent in an ABM with individual behavioral models or rules that determine how the agent behaves in response to its local virtual environment.
These behavioral models are intended to be plausible approximations of real-world individual decision-making.
Simple and intuitive behavioral rules can result in a wide range of emergent macroscale phenomena that are not intuitively predictable from the behavior of each agent~\cite{schelling1969models}.
Some simple behavioral models may produce observed behaviors that do not fit the standard rational actor model or other assumptions of utility-maximizing decision-making (and the multi-agent corollary, the Nash equilibrium).

This work explores a solution to the problem of specifying agent behavioral models that implement the agent's \emph{identified} utility using the reinforcement learning (RL) formalism.
The use of the RL formalism may also increase the range of policy research questions ABMs can be used to address.
We suggest that the RL formalism may be an efficient default abstraction for specifying behavioral models in ABMs.

\subsection{Behavioral Validity in ABMs}
The analytic promise of agent-based modeling is achievable only if the ABMs are valid representations of real-world behaviors.
The validity of an ABM depends strongly on the plausibility of the agents’ defined behavior models.
And this is especially true when ABMs are applied to inform real-world policy decision-making.
Small changes to behavioral models can sometimes lead to macro-predictions that are qualitatively very different.
And it is not always clear which behavioral model should apply since the behavioral models are not necessarily validated with data.
This is especially poignant when a particular ruleset leads to some agents behaving obviously sub-optimally and this sub-optimal behavior drives observed macro outcomes.

Several traditional approaches aim to assure the validity of ABM behavioral models.
The most direct is to base behavioral models on \emph{data-driven regression models} that relate state or environmental factors to the agents’ decision outcomes or profiles.
This approach is prominent in microsimulation modeling.
There are assumptions embedded in this practice that may undermine the validity of the approach e.g. stationarity (i.e. the patterns of behaviors in observed training data will persist) and causality (i.e. that the observational data samples sufficiently capture the causal structure of agent decision-making).

\emph{Cognitive architecture modeling} offers another approach to developing valid behavioral models.
Here, the modeler specifies a psychologically plausible model of agent decisionmaking for the ABM (sometimes informed by survey data).
Each agent then simulates a private version of the identified cognitive architecture when making decisions in the ABM environment.
Cognitive architectures have a long history in the field of artificial intelligence.
Examples include the ACT-R and Soar architectures~\cite{johnson1997control,anderson1996act}.

This approach to decision modeling aims to be more causal and mechanistic than the data-driven microsimulation approach.
But, arguably, this detailed lower-level modeling is disconnected from the level of inquiry the ABM is designed to support.
The policy analyst often cares more about the representativeness of observed agent behaviors rather than the specific mental calculi (often not even uniquely identifiable) that produce observed decisions.

Modelers need an efficient approach for specifying decision or behavioral models that is: Representative of real-world causal decision-making;  Agnostic to the exact internal structure agents’ decision-making; $\&$ Flexible or adaptable to the changes in agents’ local context over time.

\subsection{Specifying Valid Behaviors using Reinforcement Learning}
We explore the use of RL policies as behavioral models for ABMs.
In this approach, we specify or identify: what an ABM agent can \emph{observe}, what they can \emph{do}, and what they \emph{value} in an environment.
Given these pieces of information, we specify a generic-but-flexible policy model that the agents can tune using powerful learning algorithms applied to experience data gleaned from interaction with the ABM environment.
Each agent tunes its decision policy to optimize how much reward it can expect to extract from its interactions with the environment.
In essence, we implement utility-maximizing virtual agents in our ABMs with the key generalization that agents can have diverse specifications for their private utilities.
And we represent the agents’ decision functions with differentiable models (neural networks) for ease of implementation and optimization.

This approach abstracts away the agents’ internal structure of decision-making.
Each agent can continue to learn and adapt effectively even when the virtual environment is operating a completely new regime.
And, if we \emph{know} and \emph{can represent} the utility functions of real-world agents,
then our policies can be representative of real-world decision-making.
This approach could be preferable to the default assumption of ad-hoc rule-based agents in ABMs.

\subsection{Scoping Our Exploration}
We developed proof-of-concept AI-based behavioral models in multiagent decision-making contexts.
We focused on applying these models to policy-relevant ABMs.
Applicable ABMs include ABMs of the stock market, vaccination, taxation, and health insurance market behaviors.
This line of inquiry is motivated by the observation that some policy-based ABMs are similar to or simpler in game-theoretic structure to video games.
We target our efforts to explore the use of RL-based behavioral models for multiagent experiments with ABM.
The outcomes of our efforts are a set of ported ABM environments and templates for RL-based behavioral models that enable policy researchers to:
\begin{itemize}
	\item Equip the agents in ABMs with flexible capacity adaptive intelligence;
	\item Explore the effects of interaction between diverse sets of strategic behaviors (e.g. bounded rationality, collusion) in ABM populations; $\&$
	\item Possibly learn behavioral policies that better explain empirically-observed behaviors (Similar to Epstein’s discussion of inverse explanations) ~\cite{epstein2006generative}.
\end{itemize}

We show the value of our efforts by running experiments to examine the effects of RL behavioral models in two policy-relevant ABMs: the minority game $\&$ the flu vaccination game. The basic research questions we aim to answer in these experiments include:
\begin{itemize}
	\item Examining the effects of heterogeneous decision models (RL vs. default heuristics) in an ABM;
	\item Examining the effects of differing levels of agents’ intelligence and/or memory on decision-making in an ABM; $\&$
	\item Examining the capacity for populations intelligent agents to learn coordination or synchronization under conditions of limited or no direct communication.
\end{itemize}

The rest of this discussion has the following structure. The next section (\ref{sec:method}) outlines key design and modeling decisions we made in the process of developing our AI-equipped ABMs, both in the single-agent and the multiagent setups. Sections \ref{sec:mingame} $\&$ \ref{sec:flu} discuss our research findings when we implemented two examples of policy-relevant AI-equipped ABMs. Section \ref{sec:close} summarizes our findings.

\section{Methodology and Technical Details}\label{sec:method}
We developed a collection of policy-relevant multiagent environments $\&$ implementations of single and multiagent RL algorithms.
The goal was to connect both lines of development to produce policy-focused ABMs that feature RL-based decisionmaking agents.
The rest of this section explores key details in these efforts.

We identified some new and pre-existing policy-focused ABMs that are relevant for our purposes.
These include a Flu vaccination ABM~\cite{nowak2017general,vardavas2007can,vardavas2013modeling}, and a tax-evasion ABM~\cite{vardavas2019rand}.
They are listed roughly in increasing order of model complexity.
We decided to start with a simpler ABM that captures key policy-relevant behavior: the Minority Game ABM~\cite{challet2013minority,challet2001minority} which is based on the El Farol Bar problem~\cite{arthur1994complexity}.
It represents a simplified abstracted version of the stochastic game at the heart of the Flu and Tax Evasion models.
Section \ref{sec:mingame} discusses this ABM in more detail.

We encapsulate each ABMs using standardized gym environment interfaces.
This abstraction effectively separates the mechanics of the ABM from the behavioral models of the agents. Each environment wrapper implements (using Figure 2.1 as background context): $~s_{t+1},~r_{t},doneQ=env.step \left( a_{t} \right) $, $s_{0}=env.reset \left( \right) $.

\subsection{RL Behavioral Models: the Single-agent Case}
Given a goal within the virtual environment, we specify a reward function such that achieving the environment goal maximizes the agent’s reward.
We equip an agent with a neural-network-based model and statistical learning algorithms to enable the agent to adapt its behavioral policy to maximize the reward function~\cite{sutton2018reinforcement}.
In this setup, it is also possible to model complex or strategic goals like maximizing the volatility of a market or a stock price or optimizing the explanatory or predictive power of agent's policies for observed real policies~\cite{epstein2006generative}.

\begin{figure}[!ht]
\centering
		\includegraphics[width=0.9\columnwidth]{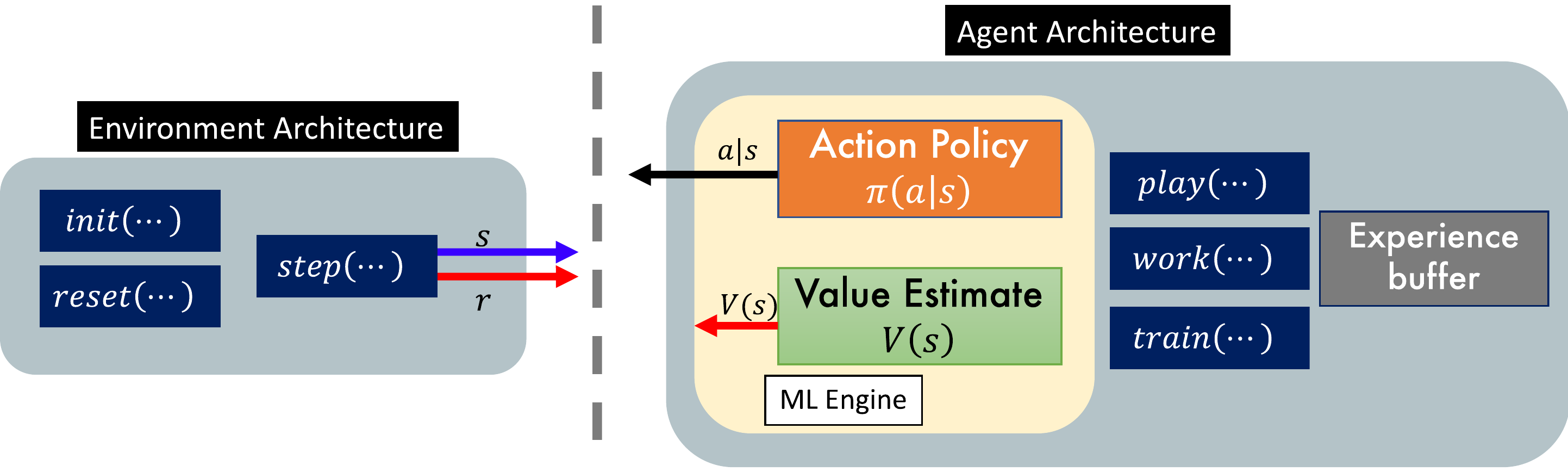}
		\caption{: Schematic of interaction components between an agent and the ABM environment (arrows extending out of boxes denote exposed signals).}
		\label{fig:schematic}
\end{figure}

Figure~\ref{fig:schematic} depicts key components and interactions in the environment and agent modules.
The agents implement: a behavioral policy model $a_{t}=agent.act \left( s_{t} \right)$, an experience buffer, and \emph{play/work/train} functions.

\vspace{-4pt}
\vspace{-4pt}

The basic template for the agent's decision models, the \emph{embodied\footnote{
	The name is a nod towards the concept of embodied cognition wherein an agent’s cognition is strongly influenced or determined by how the agent senses, reasons about, and acts in its surrounding environment~\cite{lakoff2012explaining}.
} architecture} is modular, based on a basic actor-critic architecture~\cite{sutton2000policy}.
We have flexibility in what kinds of learning algorithms we can apply to this basic architecture. And we can always augment the architecture with auxiliary intelligence artifacts (e.g. see central Q-function in the MAC algorithm, Table 4).


We default to using policy-gradient adaptation algorithms on actor-critic-style models for our agent behavioral models.
This approach is more flexible for representing mixed-strategy policies.
And that flexibility in policy structure will be crucial in the multiagent RL (MARL) context.

\subsection{RL Behavioral Models: The Multiagent Case}
The Multi-agent RL (MARL) context has more complexity than the single-agent control problem described so far.
We need to make adaptations to the learning models and algorithms to account for multi-agency~\cite{stone2000multiagent,tuyls2012multiagent,bucsoniu2010multi}.
MARL algorithms are often generalizations of single-agent RL algorithms~\cite{bucsoniu2010multi,stone2000multiagent}.
But this generalization strategy is inadequate because the convergence of the single-agent control algorithms described assumes that the agent is interacting with a somewhat \emph{stationary} environment. Multi-agent contexts will tend to violate stationarity~\cite{lowe2017multi}.

Furthermore, the optimal strategy for one member of a population depends on the current strategies of the rest of the population. We may need to replace the concept of an independent optimal strategy with the concept of an evolutionarily stable strategy (ESS)~\cite{smith1973logic,axelrod1981evolution} instead.

The violation of basic assumptions (stationarity, the Markov frame, independence of agent rewards, the existence of stable unconditional policy targets) can be severe enough to render traditionally powerful learning algorithms unstable in multiagent environments~\cite{lowe2017multi,bucsoniu2010multi}.
We implement a Multi-agent Actor-Critic (MAC) algorithm based on~\cite{lowe2017multi} to address the non-stationarity issue in our models.
The MAC algorithm furnishes each agent with access to a central global critic, but only during training.
Our agents revert to independent adaptation after deployment.

Examples may help illustrate some of key MARL-specific concerns in more concrete terms.
For example, competing firms must consider the rigidities in the pricing of other firms and how adaptive other firms are in changes to the market.
As another example, computer network defense teams must consider the sophistication and tools available to the adversaries, especially knowing that the adversaries are actively trying to evade detection.

We model the multiagent environment in a couple of ways.
The first is the simplest setup: the independent population model~\cite{lanctot2017unified}.
Each RL-driven agent in the population has a standalone embodied model with which they interact with the shared environment.
In this model, there is no internal communication or direct private transfer of information among agents.
Intra-agent communication is still possible via \emph{stigmergy}~\cite{theraulaz1999brief} if the observed state variables are sufficiently detailed.
This setup is good for modeling fully competitive games with no explicit collusion allowed.
But this independent setup is often imperfect, both as a model of real-world behaviors and for robust multiagent learning dynamics~\cite{lanctot2017unified}.

The second setup is to enable some amount of communication and coordination across the agent population.
Communication can be in the form of sharing state information, sharing private decision propensities, or network weight-sharing.


\section{Experiment 1: RL Behavioral Models in the Minority Game}\label{sec:mingame}
We deploy RL models in the classic ABM known as the minority game.
The minority game is a classic environment for studying adaptive and learning agents.
The game generalizes the El Farol Bar problem.
Specifically, $N$ agents must decide whether to go to the El Farol bar or to stay home.
If more than half of the agents go to the bar, then the bar is too crowded, and the agent would be better off staying home.
On the other hand, if less than half of the agents go to the bar, then the best action is to go to the bar because it is lively and not too crowded.
Thus, the minority game is named as such since each agent prefers to be in the minority group.

The minority game has been extensively studied and prior work provides us with a benchmark into how well the RL agents are performing.
This is important because it is often difficult to know when an RL algorithm converges to a ``good''  local optimum.
Prior results in the minority game will serve as a guidepost.
Our minority game implementation parallels the literature~\cite{manuca2000structure} with the exception that we enable the use of an RL-based behavioral model for one or more agents in the population of $N$ players.

\subsection{The Default Agent Behavioral Model}
In each time step, agents make a binary decision, which we represent by choosing either $0$ or $1$. At the end of the time step, the agents observe the minority and majority group. That is, the agents observe whether there were more agents that chose $1$ or more agents that chose $0$. Crucially, the agents do not observe \emph{how many} other agents choose $0$ and $1$ but only observe which choice was selected by less than half of the agents, which we call the minority group.

Each agent is endowed with a \textbf{memory} parameter, m, that determines how many past realizations of the game the agent considers when making its next decision. For example, if m=3, the agent bases its binary decision at time $t$ on its observations in time $t-1$, $t-2$, and $t-3$. A \textbf{strategy} is a mapping from  $2^{m}$  to $\{0,1\}$. That is, a strategy specifies an action for all possible length-\emph{m} memories. For example, if m=2, a strategy would specify an action for each of the possible memories in $ \{ $ $ \{ $ 0,0$ \} $ , $ \{ $ 0,1$ \} $ , $ \{ $ 1,0$ \} $ , $ \{ $ 1,1$ \} $ $ \} $ .

At the start of the game, each agent draws $k$ random strategies with replacement. Each agent is seeded with an initial strategy. Also, a random initial history is drawn uniform randomly. At each timestep, an agent selects an action specified by the strategy with the most accumulated points. A strategy accumulates a point in each timestep if the action specified by the strategy would have put the agent in the minority group.

\subsection{RL-based Behavioral Models}
In the model, the RL agents do not choose their strategy like the basic agents. Instead, the RL-based agents choose their actions based on inference via tuned neural networks. To control for the RL agent’s memory, we change the number of inputs into the neural network. For example, if we want to model the case where the RL agent chooses its action conditional on the previous 3 minority groups, then the input to the deep neural network is a three-dimensional binary vector. Later, we discuss the possibility of modeling an agent’s memory with recurrent neural networks. We train the RL agents in response to its experience in the game using a variety of policy gradient algorithms (REINFORCE, REINFORCE with baseline, $\&$ Actor-Critic).

\subsubsection{Single-Agent Experiments}
We train $1$ reinforcement learner to play against $N-1$ basic agents.
In experiments, N=301, and each basic agent has 2 strategies with a memory of $2$.
The number of agents in each group is periodic and follows a perfectly predictable pattern.
This implies that the winning group is also periodic and follows the pattern $'1,1,1,0,0,0,1,0.'$
Therefore, with sufficient memory, a reinforcement learner should be able to perfectly predict the minority group simply by observing the history of winners.

\begin{figure}[htb]
	 \centering
		\includegraphics[width=0.75\columnwidth, keepaspectratio]{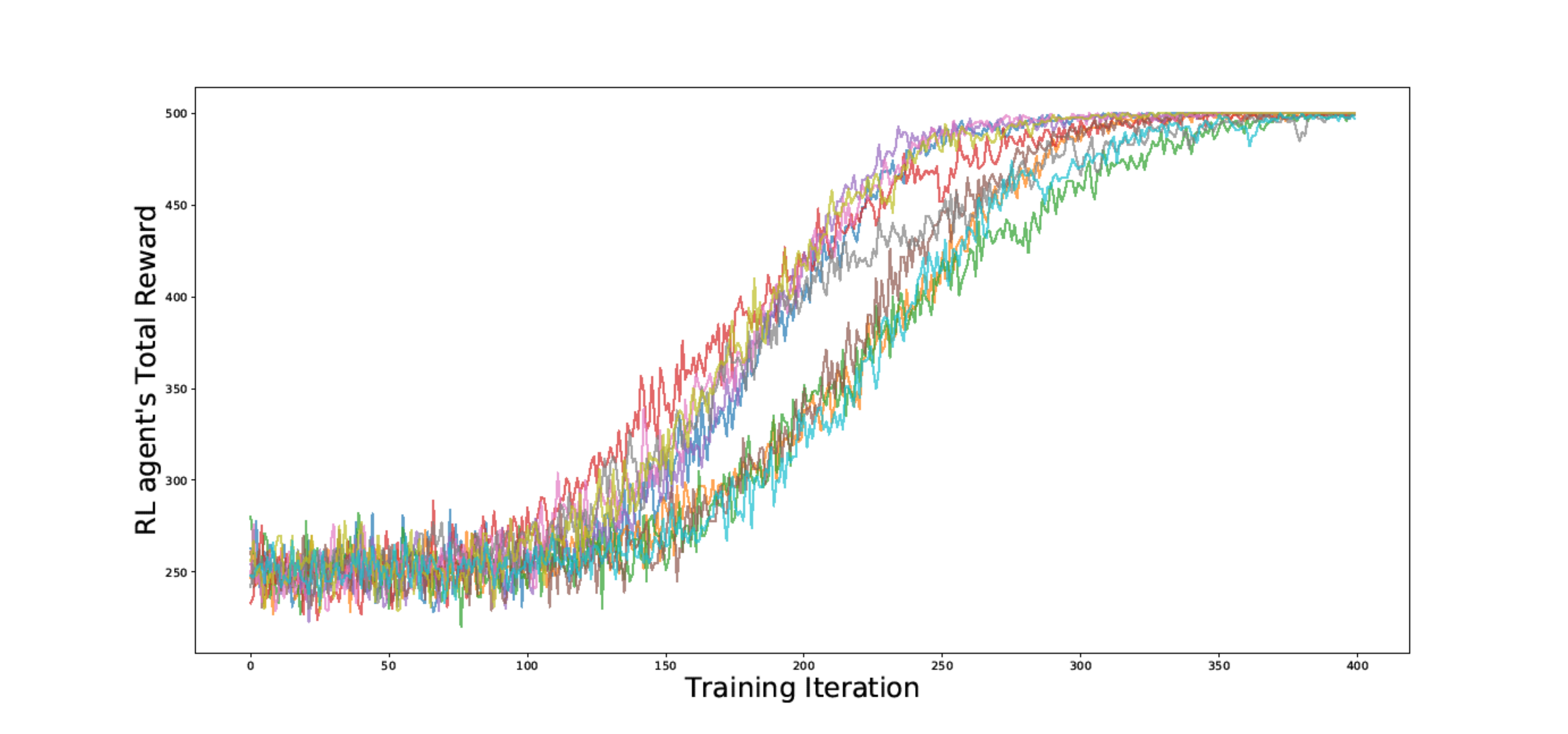}
		\caption{Time series of attendees for 500 time steps, 301 agents each with 2 strategies and a memory of 2}
		\label{fg:tseries}
\end{figure}

To test this, we train a policy gradient neural network with 4 hidden layers and 20 nodes per layer in the minority game environment.
To aid in computation, the RL agent's look-ahead is limited to $5$ time-steps.
The input to the neural network is the last 3 winning groups, which is one greater than the basic agent's memory.
Figure~\ref{fg:tseries} shows that the RL agent does indeed learn the optimal strategy and correctly guesses the minority group each time.

A crucial feature of this experiment is that for each of the 10 trials, the RL agent is trained against a different set of agents but the set of agents stays constant within each of the 10 trials.  In other words, in the first trial, we sample a set of 300 basic agents and train a neural network against those 300 agents using 400 training steps (recall there is randomness in the agents when drawing their strategies).  In the second trial, we sample a different set of 300 agents and retrain.  Therefore, Figure 3.2 shows that an RL agent plays optimally against the agents it trained against.  However, a key question regarding generalization is then how well does an RL learner perform against agents it did not train against?

\begin{figure}[ht]
	 \centering
		\includegraphics[width=0.75\columnwidth, keepaspectratio]{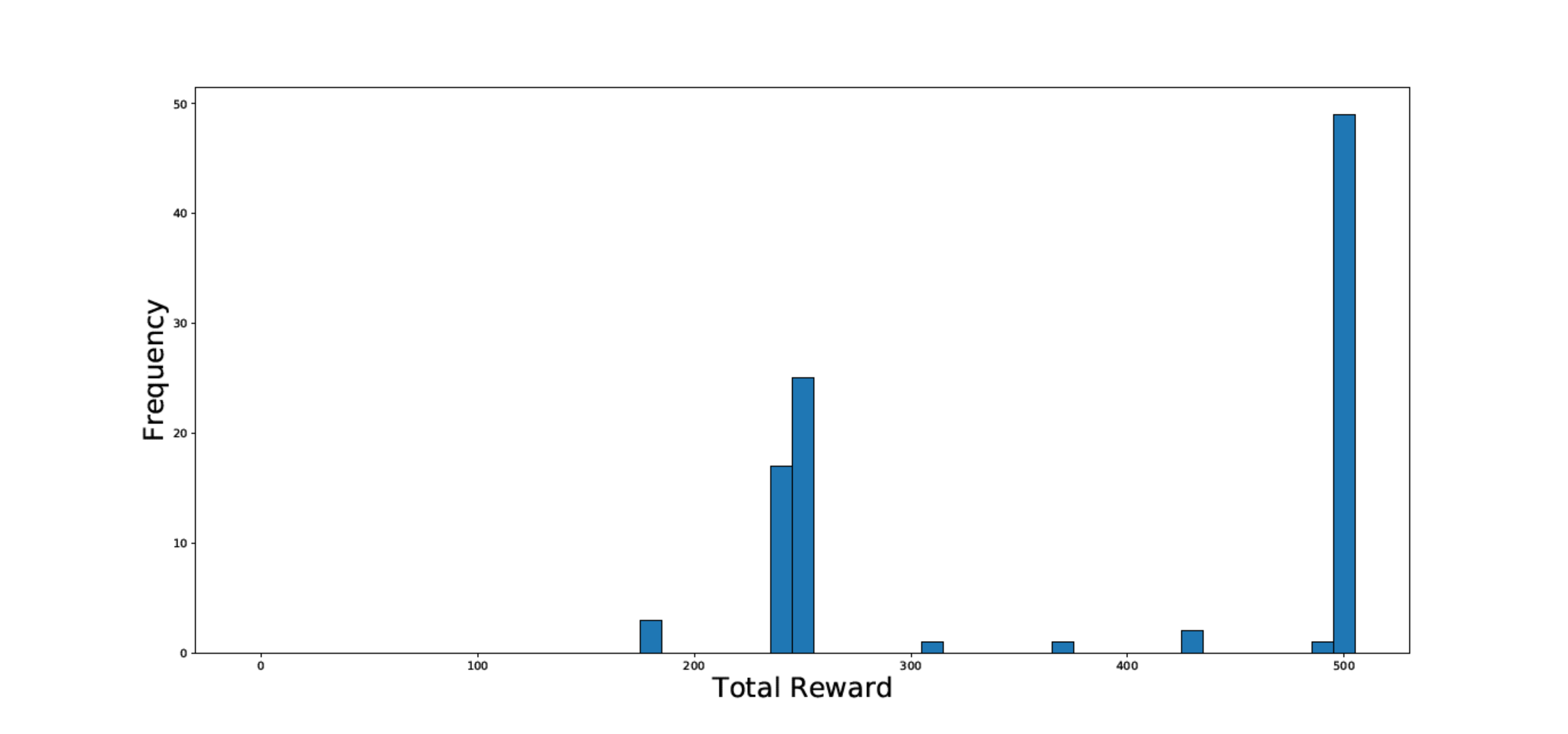}
		\caption{  A trained RL learner playing against 100 different randomly drawn populations}
		\label{fg:randpop}
\end{figure}

Figure~\ref{fg:randpop} shows that the RL learner's behavior is not always optimal against a different randomly drawn population.
This is especially surprising since every randomly drawn population with 2 strategies and a 2 time-step memory yield periodic and predictable patterns.
However, the key point is that the precise pattern depends on the labeling of the groups and the RL agent is sensitive to such labeling.
In other words, the RL algorithm is not robust to relabeling.

While an agent’s performance trained on one population does not generalize, the next obvious question is: can an agent with enough memory to play perfectly against any given population be trained to play perfectly against all populations?  To answer this, we once again train an RL agent with the same parameters.  However, in this experiment, we resample basic agents for each training iteration.  In other words, the RL learner plays against a different population for every training iteration.

\begin{figure}[ht]
	 \centering
		\includegraphics[width=0.75\columnwidth, keepaspectratio]{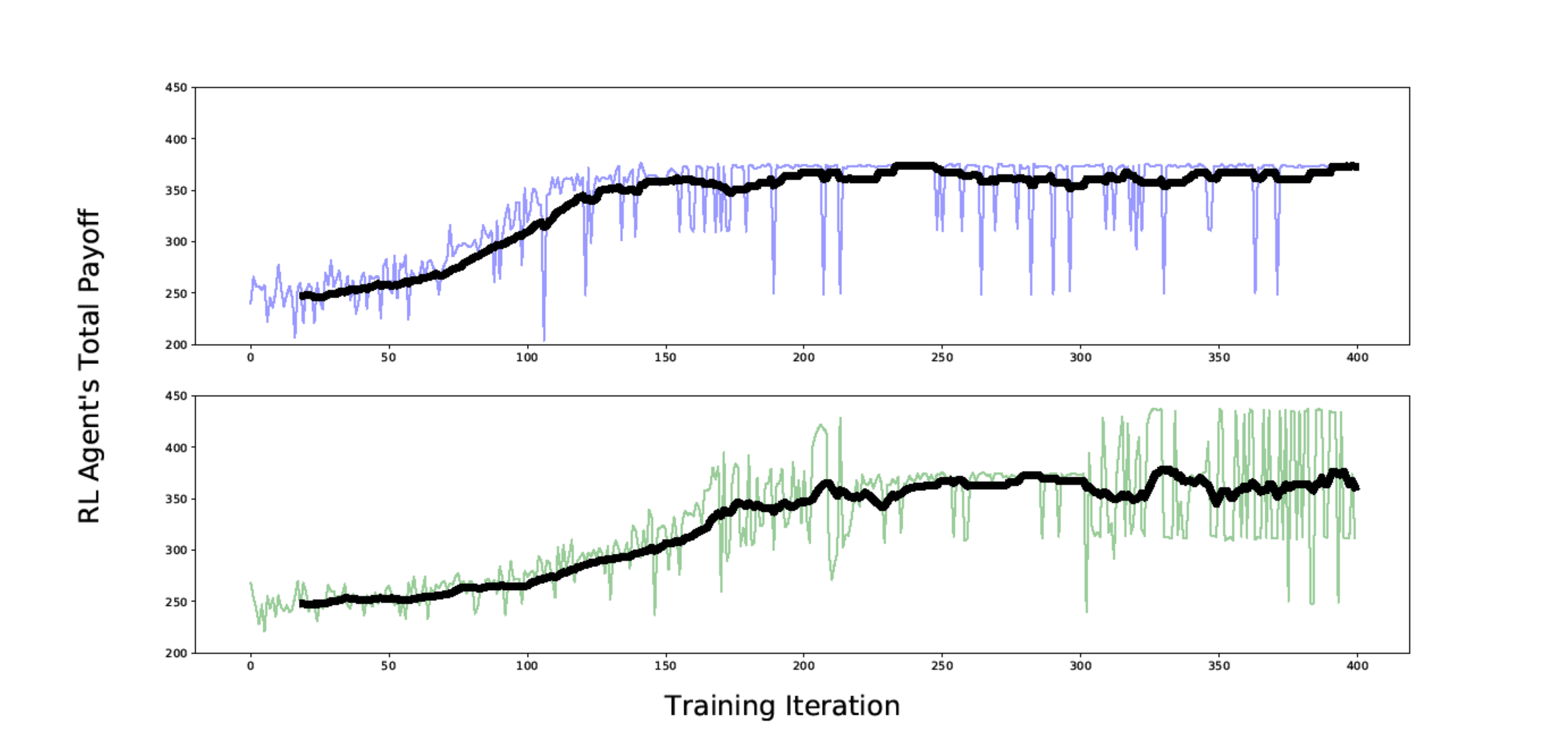}
		\caption{Two different rounds of training with a neural network with 3 time steps of memory. In each round, there were 400 training steps with an episode length of 500 time steps. The black line is a rolling mean with a window length of 20.}
		\label{fg:twomg_tst}
\end{figure}

\begin{figure}[htb]
	 \centering
		\includegraphics[width=0.75\columnwidth, keepaspectratio]{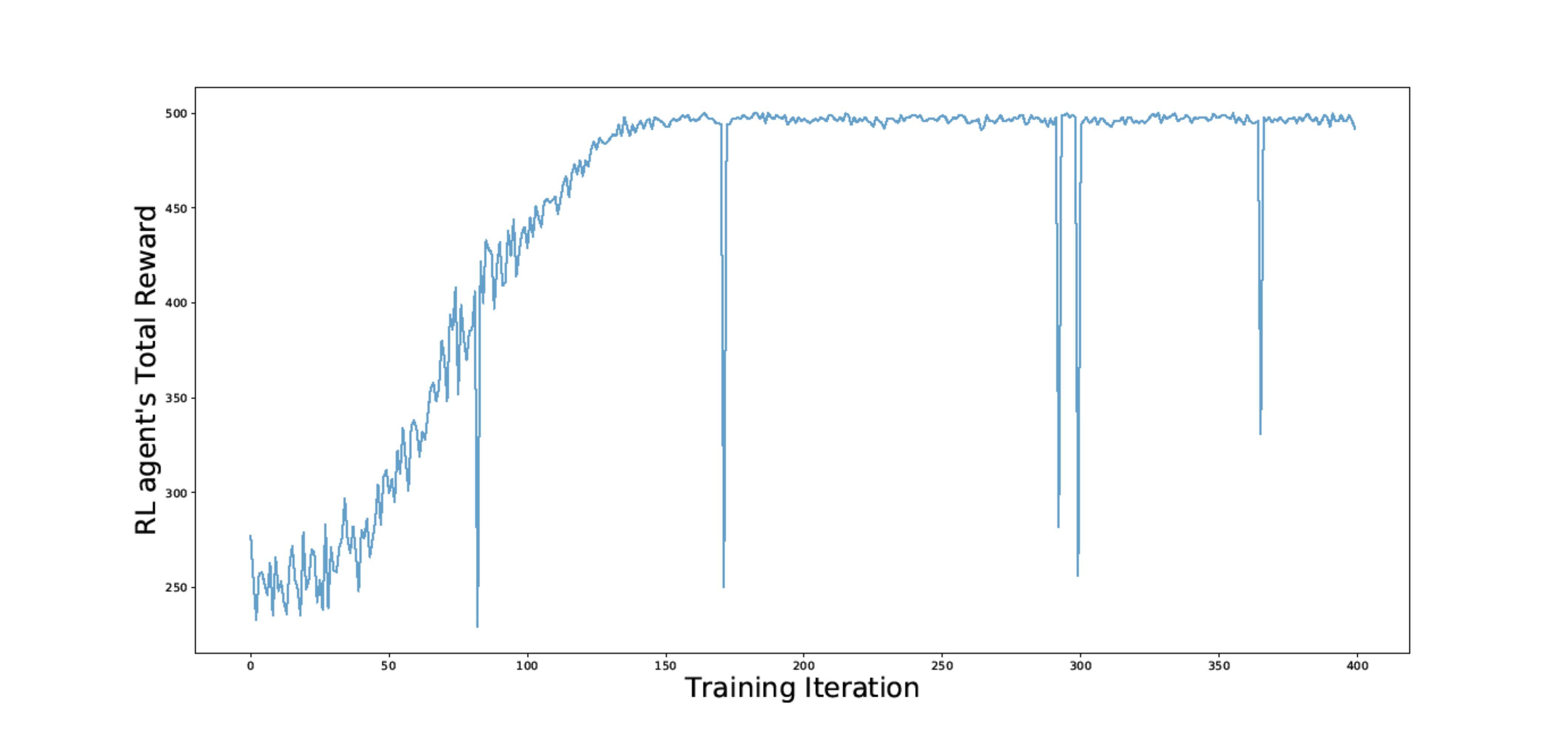}
		\caption{Training with an RL agent with 5 time steps of memory. In each round, there were 400 training steps with an episode length of 500 time steps. The low spikes are the case where the initial distribution of agents yields ties.}
		\label{fg:5steps}
\end{figure}

\begin{table*}[ht] 
	\centering
	\begin{tabular}{p{0.5\textwidth}p{0.5\textwidth}}
		\multicolumn{1}{p{0.475\textwidth}}{
				\includegraphics[width=0.45\textwidth,keepaspectratio]{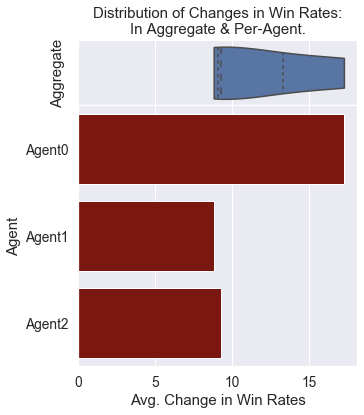}
		} &
		\multicolumn{1}{|p{0.475\textwidth}}{
				\includegraphics[width=0.45\textwidth,keepaspectratio]{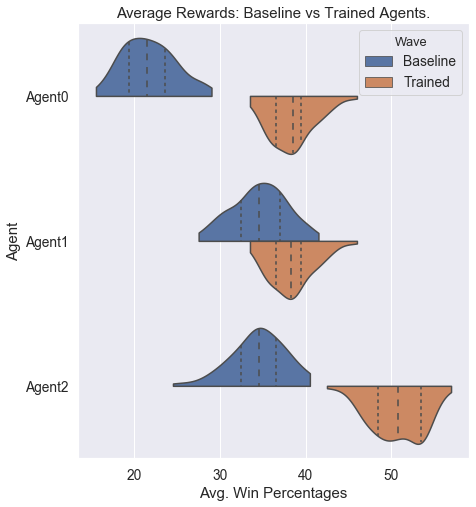}
		} 
	\end{tabular}
\caption{Improvements from training three minority game players (out of a population ten player) using RL-based behavioral models. The remaining 7 players act according to the default behavioral model for the minority game.}
\label{fg:mgmres}
\end{table*}

Figure~\ref{fg:twomg_tst} gives two examples of how the RL agent learns when it faces a different population of basic agents for each training iteration.
The key insight is that the RL agent is \emph{not} able to correctly choose the minority group every time, even though for a given population of basic agents, the sequence of minority groups is deterministic.
This is because 3 observations in the past are not sufficient to determine which population the learning agent is playing against.
Therefore, since the RL agent cannot condition its action on the full periodic sequence but can only use the last three observations, it cannot always correctly choose the minority group.

Stated succinctly, an RL agent that has enough memory to play perfectly against any given draw of an initial distribution is not sufficient to play perfectly against all draws of the initial distribution.

Since $3$ units of memory are insufficient for the RL learner to predict the minority group in all cases, we extend the memory parameter.
It can be shown that the RL learner cannot learn perfectly against all distributions of agents until it has a memory of length $6$.
Figure~\ref{fg:5steps} shows that this is indeed the case.
The troughs in the figure represent cases where the initial distribution of agents yields ties in the minority group, which changes the dynamics of the system.
However, the probability of this occurring goes to zero as the number of agents gets larger and thus can safely be ignored.

\subsubsection{Multiagent Experiments}
We also explore some of the dynamics of the game when \emph{more than one agent} in the population uses RL-based behavioral models. All agents interact in a shared minority game. And all agents make decisions based on observable information (in this case, a fully shared history of recent winning choices). Agents using the default behavioral model continue to make decisions based on their adaptive strategy books. RL agents make decisions based on the outcome of their stochastic policy neural networks.
The reward signal in our setup is simply a binary signal $(0/1)$ corresponding to \emph{individual} loss or win. We assume no explicit collusion or direct communication within the RL agent sub-population during the test phase of the experiments.

Preliminary results (Table~\ref{fg:mgmres}) show that the multiagent RL sub-population is able to adapt to improve both its aggregate and individual rewards in the minority game. Figure 3.6 shows learning improvements in reaped rewards for an RL subpopulation of 3 agents playing in an overall population of size N=10. The game is structured with a standard memory of length m=3 and 4 strategies per agent in the default behavior model. The second panel of the figure compares the agents’ distribution of rewards pre- and post- RL training (601 training epochs using the MAC algorithm; see Table 4 in the Appendix). The primary takeaway from our experiments is that the RL training algorithms are able to learn \emph{reward-seeking behavior}. However, experiments comparing the performance of the multiagent RL behavioral model to the default behavioral model yield equivocal results.

\section{Experiment 2: Reinforcement Learning in Flu Propagation and Vaccination on a Social Network}\label{sec:flu}
In experiment 2 we use and modify an existing ABM that models the complex dynamical interplay between vaccination behavior, influenza epidemiology, and influenza prevention policies over a synthetic population that is representative of the population of Portland, OR. Our experiment replaces individual agents in the model with agents that use a different behavioral algorithm based on standard RL rules. This ABM builds off from previous work~\cite{vardavas2013modeling,nowak2017general,vardavas2007can}.
These prior works contain high-level descriptions of the flu vaccine ABM we adapt here. And further details of the complete Flu transmission ABM model is provided as supplementary material.

Central to the ABM's dynamics is the assumption that personal and social-network experiences from past influenza seasons affect current decisions to get vaccinated, and thus influence the course of an epidemic. The model proceeds iteratively from one flu season to the next. Agents in the model represent individuals who determine whether or not they get vaccinated for the present season.
Their collective decisions drive influenza epidemiology that, in turn, affects future behavior and decisions.

The ABM environment is described by: (A) the behavioral model; (B) the social and contact network structures; and (C) the influenza transmission dynamics.
The model simulates within-year influenza transmission dynamics on the network using a Susceptible-Infected-Recovered (SIR) model.
The full model also considers additional complications and confounding factors such as whether the agents are infected with other non-influenza influenza-like illnesses (niILI).

The ABM's underlying network structure was informed by generating a reduced but statistically representative network with roughly ten thousand individuals taken from open datasets from the Network Dynamics and Simulation Science Laboratory (NDSSL).
This dataset represents a synthetic population of the city of Portland, OR and its surroundings in an instance of a time-varying social contact network for a ‘normal’ day, derived from the daily activities.

\subsection{The Default Agent Behavioral Model}
The behavioral model specifies the way that agents belonging to the different outcome groups put different emphasis/weights on their personal experience and local and global feedback information and how they evaluate their choice.
Agents are assumed to remember experiences from past years, and these determine how they change their vaccination behavior.
Therefore, agents update their propensity to get vaccinated for the following season based on a weighted sum of their most recent evaluation and evaluations made in past years (discounted over time).

The ABM’s default behavioral model was informed from different data sources.
For example, to inform the behavioral model and quantify its relevant parameters, we use internet surveys on a random subsample of a longitudinal panel survey of a nationally-representative cross-section of the US.
Specifically, the surveys were used to quantify how individuals’ behavior towards vaccination changes based on key flu-related experiences and beliefs.

The behavioral model assumes that at the beginning of year $n$ , an adult individual $i$ will get vaccinated with probability  $w_{n}^{ \left( i \right) }$  given by a convex-combination expression
\begin{align}
	w_{n}^{ \left( i \right) }= \beta _{a} \upsilon _{n}^{ \left( i \right) }+ \beta _{b} \phi _{n}^{ \left( i \right) }+ \beta _{c}. \psi ^{ \left( i \right) } ,
\end{align}
where the  $ \beta $  values are convex coefficients that weight a number of factors that influence each agent’s vaccination probability. Vaccination probability  $w_{n}^{ \left( i \right) }$  is modeled as a Bernoulli mixture model with contributions from the following three factors:
\vspace{-3pt}
\begin{itemize}
	\item  $ \upsilon _{n}^{ \left( i \right) }$ : the agent’s \emph{intrinsic} adaptation and learning. This includes signal contributions from its social network. And this factor represents a weighted sum of past experiences with flu infections and vaccination. The weighting scheme lends more importance to the more recent experiences.
	\item  $ \phi _{n}^{ \left( i \right) }$ : the influence of healthcare workers’ (HCWs) recommendations on the agent’s probability of getting vaccinated. $\&$
	\item  $ \psi ^{ \left( i \right) }$ : a set of socio-economic, biological and attitudinal factors (e.g., cost, medical predisposition and convenience of obtaining vaccination) that are assumed to be stationary and relevant to vaccination decisions.
\end{itemize}


The full ABM considers all three factors. However, for this model, we decided to use a simplified version of the behavioral model which considers a homogeneous population that makes their vaccination decisions exclusively based on the present and past evaluations of their vaccination decisions and outcomes. As such in this simplified model we only consider the agent’s intrinsic adaptation and learning factor, and set the value  $\beta_{\alpha}$  to 1. Hence, agents are not influenced by recommendations made by their physicians. Moreover, the simplified model that we are considering here also removes other relevant details.

\subsection{RL-based Behavioral Models in Influenza ABM Simulations}
For all RL experiments on this model, the agents gain rewards by not being infected during a flu season. They observe their prior-year flu outcomes and healthcare worker recommendations. They act by either choosing vaccination or not.

\begin{figure}[htb]
	  \centering
		\includegraphics[width=0.5\columnwidth, keepaspectratio]{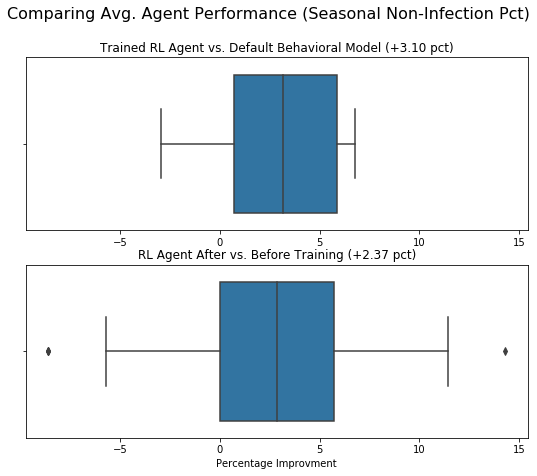} 
		\caption{Comparing the performance of an RL agent in the Flu ABM environment. The RL model a.) is able to improve its outcomes after multiple training epochs (bottom panel), $\&$  b.) outperforms the default behavioral model on average (n=100 seasons) by a couple of percentage points (top panel).}
		\label{fg:flu_rl_dbm}
\end{figure}

\subsubsection{Single Agent Learning in the Flu Environment}
Before running our experiments in the ABM of influenza vaccination, we ran the simulation model with no RL agents until the dynamics of aggregated macro quantity reached a stationary state. In this state, our model has reached a dynamic equilibrium of detailed balance analogous to kinetic theory. At this point, we sample 1 or N agents and change their behavioral model to that specified by the RL algorithm. Figure~\ref{fg:flu_rl_dbm} below shows the performance of the RL model after $750$ training iterations (using a basic actor-critic learning model). The reward signal in this game is the number of seasons the agent goes through without contracting the flu. The basic finding from this experiment is that single-agent reward-driven learning is effective in this Flu ABM.

\subsubsection{Multiagent Learning: Effects of Agent Degree \& Action Synchronization}
We run a set of experiments to examine how much influence an agent’s degree centrality has on the effectiveness of its learning. We draw two separate subpopulations at random each of size $n=40$. The first sub-population in the lower quartile of network degree centrality in the overall synthetic population. The second subpopulation is in the top quartile of network degree centrality. Both sub-populations are trained with the same algorithm (MAC) starting from identical policy templates.
Then we examine how much the ensembles improved after about $1500$ learning iterations.

Table~\ref{fg:flum_res} shows their improvements. The more basic result from this experiment is that multiagent reward-driven learning is effective in this Flu ABM. The results also suggest preliminary evidence in support of the hypothesis that low-degree centrality agents are more effective at learning to avoid the flu compared to high-degree centrality agents. This aligns with our intuitions since higher degree nodes have more contact and thus more paths for infection. But the experiment needs more data to test the hypothesis further.

We also explored the question of synchronization in this experiment. We are concerned with the emergence of synchronized behavior in ensembles on non-communicating learning agents. We examine the evidence for this by calculating an average correlation matrix for the observed actions from an ensemble of trained agents. Any synchronization (positive or negative) between agents will be exposed by high magnitude correlation matrix entries or bands.
No synchronization was evident in this experiment (and others) since the correlations are capped in magnitude at about $10\%$.

\section{Conclusions}\label{sec:close}

Our most basic finding is that single agents, playing in either of the test policy ABMs, exhibit reward-seeking behavior, and can even learn optimal policies.
There are caveats including issues with generalization and policy model memory or capacity (which may be addressable using recurrent networks).

The multiagent experiments
on both ABMs again support our motivating hypothesis that RL behavioral model can reproduce adaptive reward-seeking behavior even in a complex ABM like the flu model.
And the RL subpopulations can significantly outperform subpopulations that deployed the default behavioral model.
One observation was that naïve MARL algorithms are not as effective at learning good behaviors.
This is in line with the current MARL literature.
We adapt the MAC algorithm to address pathologies specific to multiagent games.

The RL-based behavioral models can also compare favorably with default behavioral models in ABMs.
The experimental results on this point are not as unequivocal.
The effect sizes in these experiments are smaller.
This is likely because the outcome of such comparisons is rather path-dependent and sensitive to initial conditions.

These results suggest that the RL formalism can be an efficient default abstraction for behavioral models in ABMs. The RL formalism cleanly and efficiently represents reward-seeking behavior in intelligent agents. Furthermore, the RL formalism allows modelers to specify behavioral models in a way that is agnostic to the agent's detailed internal structure of decision-making, which is often unidentifiable.

Future work would explore more complex experiments on these ABMs; explore new ABMs; and develop relevant algorithms for behavioral adaptation in these ABMs (e.g. neuro-evolutionary learning algorithms~\cite{stanley2002evolving}).

\begin{table}[htb]
 			\centering
\begin{tabular}{p{0.5\columnwidth}p{0.5\columnwidth}}
\multicolumn{1}{p{0.45\columnwidth}}{
		\includegraphics[width=0.4\columnwidth,keepaspectratio]{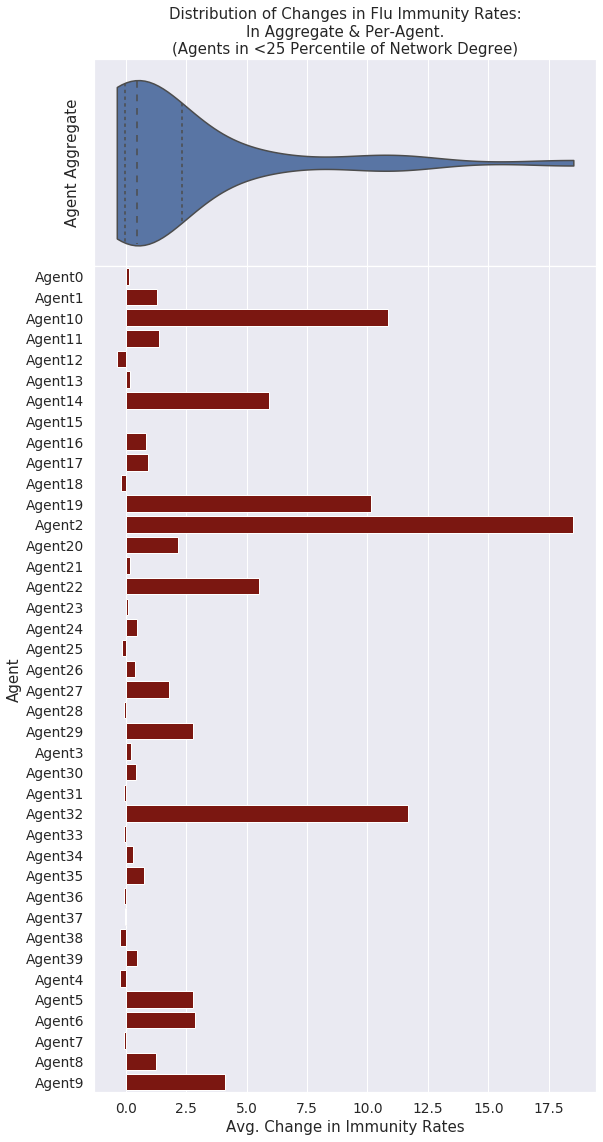}
} &
\multicolumn{1}{|p{0.45\columnwidth}}{
		\includegraphics[width=0.4\columnwidth,keepaspectratio]{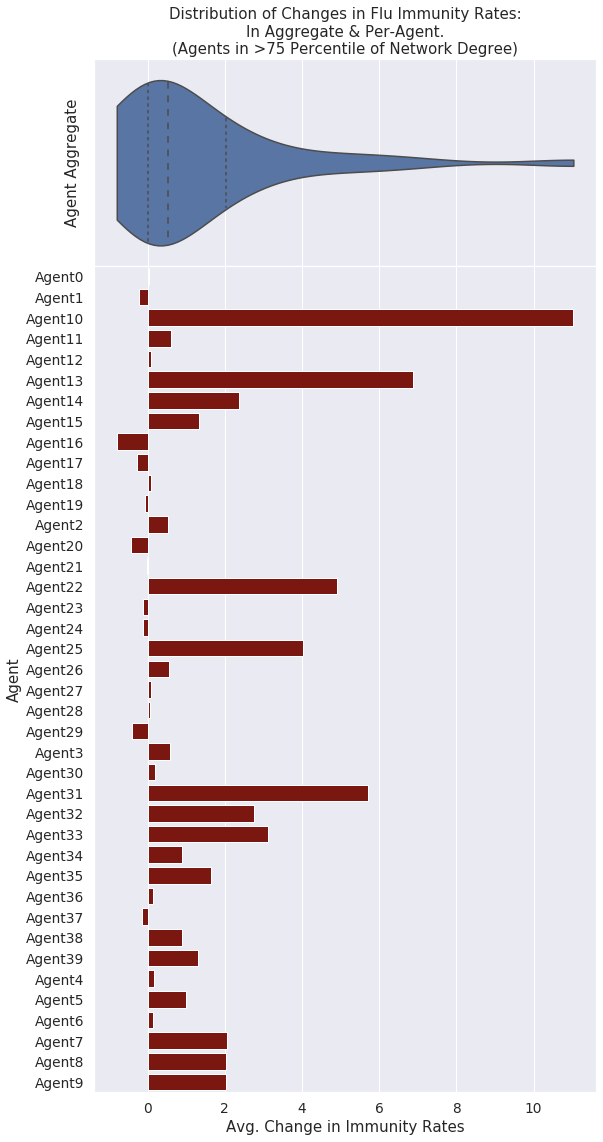}
} \\
\end{tabular}\caption{Outcome Improvements in Flu Avoidance Post-RL-training for agent ensembles in low and high quartiles of degree centrality}
\label{fg:flum_res}
\end{table}

\clearpage

\newpage


\section*{Appendix A: The Influenza Transmission Environment}
In experiment 2, we modified and use an existing ABM that models the complex dynamical interplay between vaccination behavior, influenza epidemiology, and influenza prevention policies over a synthetic population that is representative of the population of Portland, OR.
Our experiment replaces individual agents in the model with agents that use a different behavioral algorithm based on standard RL rules.
This ABM builds off from previous work~\cite{nowak2017general,vardavas2007can}.
And it is also being extended by an NIH grant research.

\begin{figure}[htb]
	  \centering
		\includegraphics[width=0.65\columnwidth, keepaspectratio]{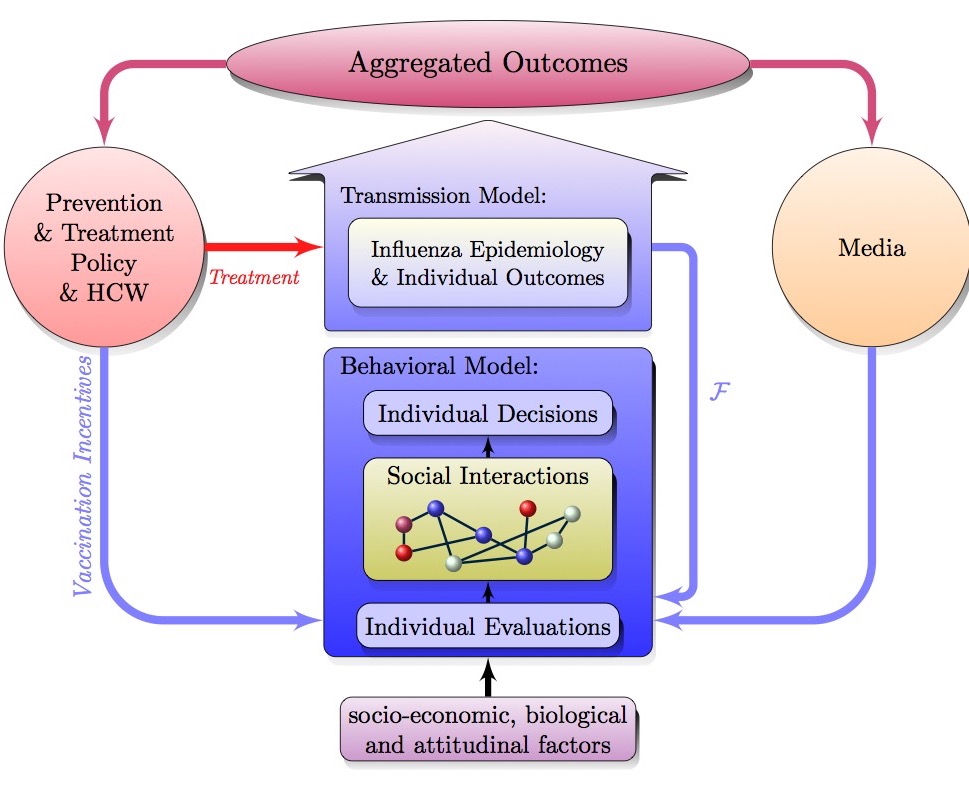}
		\caption{: Architecture of Influenza Vaccination Model.}
		\label{fg:fluabmarch}
\end{figure}

Figure~\ref{fg:fluabmarch} gives a high-level view of the ABM. Central to the ABM is the assumption that personal and social-network experiences from past influenza seasons affect current decisions to get vaccinated, and thus influence the course of an epidemic. In more detail, the model proceeds iteratively from one flu season to the next. Agents in the model represent individuals who determine whether or not they get vaccinated for the present season. Depending on their choice, the efficacy of the vaccine and the disease transmission dynamics, by the end of the season agents have either been infected or not. The model also considers additional complications and confounding factors such as whether the agents are infected with other non-influenza influenza-like illnesses (niILI), leading them to believe that they might have caught influenza. The agent population can be classified into four outcome groups:

Agents in different groups evaluate the outcomes of their choice in different ways using their personal experience with the flu and the vaccine (i.e., whether they got vaccinated and whether they were infected, including close loved ones) and using both local and global feedback information. This information includes the proportion of individuals in their social network (i.e., local) and in the population (i.e., global) that they perceive as vaccinated and infected. Our behavioral model specifies the way that agents belonging to the different outcome groups put different emphasis/weights on their personal experience and local and global feedback information and how they evaluate their choice. Agents are assumed to remember experiences from past years, and these determine how they change their vaccination behavior. Therefore, agents update their propensity to get vaccinated for the following season based on a weighted sum of their most recent evaluation and evaluations made in past years (discounted over time).

The ABM’s default behavioral model was informed from different data sources. First, to inform the behavioral model and quantify its relevant parameter, we use internet surveys on a random subsample of the RAND Corporation’s American Life Panel (ALP), a longitudinal panel study of a nationally-representative cross-section of the US was used. Specifically, the surveys were used to quantify how individuals’ behavior towards vaccination changes based on:

\begin{enumerate}
	\item personal past experiences with catching the flu and getting vaccinated;
	\item local effects due to observed experiences in the social network; and
	\item global effects based on the perceived prevalence of the flu and vaccination in the population.
\end{enumerate}

The ABM depends on an underlying network structure used to model both how influenza transmission occurs as well as how vaccination behavior influences spread. This network structure was informed by generating a reduced but statistically representative network with roughly ten thousand individuals taken from open datasets from the Network Dynamics and Simulation Science Laboratory (NDSSL). This dataset represents a synthetic population of the city of Portland, OR and its surroundings in an instance of a time-varying social contact network for a ‘normal’ day, derived from the daily activities.

\subsection*{Describing the Influenza ABM Environment}
The ABM considers a population consisting of $N$ individuals on a social network that every year considers protecting themselves, and their dependent family members from seasonal influenza. In particular, they make decisions as to whether or not to get vaccinated against seasonal influenza. Their collective decisions drive influenza epidemiology that, in turn, affects future behavior and decisions. The population is stratified according to other relevant demographic and socio-economic characteristics and are connected by an overlaying social network structure. The ABM environment is described by: (A) the behavioral model; (B) the social and contact network structures; and (C) the influenza transmission dynamics. The model proceeds iteratively as follows. At the beginning of each influenza season, every adult individual decides whether or not to get vaccinated against the flu depending on both their recent experiences with vaccination and that shared by their alters on the social network.  The model then simulates influenza transmission dynamics on the network using a Susceptible-Infected-Recovered  (SIR) model. As defined by the transmission model, an epidemic occurs every season depending on the achieved vaccination coverage (i.e., the proportion of individuals vaccinated) and whether infections span-throughout (i.e., percolate) the network. At the end of the influenza season, individuals evaluate their new experiences and accordingly modify their vaccination probabilities for the subsequent year.

Figure~\ref{fg:archBM} below shows a schematic view of all the components in the default behavioral model. It assumes that at the beginning of year  $n$ , an adult individual  $i$  will get vaccinated with probability  $w_{n}^{ \left( i \right) }$  given by a convex-combination expression

$$w_{n}^{ \left( i \right) }= \beta _{a} \upsilon _{n}^{ \left( i \right) }+ \beta _{b} \phi _{n}^{ \left( i \right) }+ \beta _{c}. \psi ^{ \left( i \right) },$$

Where the  $ \beta $  values are convex coefficients that weight a number of factors that influence each agent’s vaccination probability. Vaccination probability  $w_{n}^{ \left( i \right) }$  is modeled as a Bernoulli mixture model with contributions from the following three factors:

\begin{enumerate}
	\item  $ \upsilon _{n}^{ \left( i \right) }$ : the agent’s \textit{intrinsic} adaptation and learning. This includes signal contributions from its social network. And this factor represents a weighted sum of past experiences with flu infections and vaccination. The weighting scheme lends more importance to the more recent experiences.

	\item  $ \phi _{n}^{ \left( i \right) }$ : the influence of healthcare workers’ (HCWs) recommendations on the agent’s probability of getting vaccinated. $\&$

	\item  $ \psi ^{ \left( i \right) }$ : a set of socio-economic, biological and attitudinal factors (e.g., cost, medical predisposition and convenience of obtaining vaccination) that are assumed to be stationary and relevant to vaccination decisions.
\end{enumerate}

\begin{figure}[htb]
	  \centering
		\includegraphics[width=0.6\textwidth, keepaspectratio]{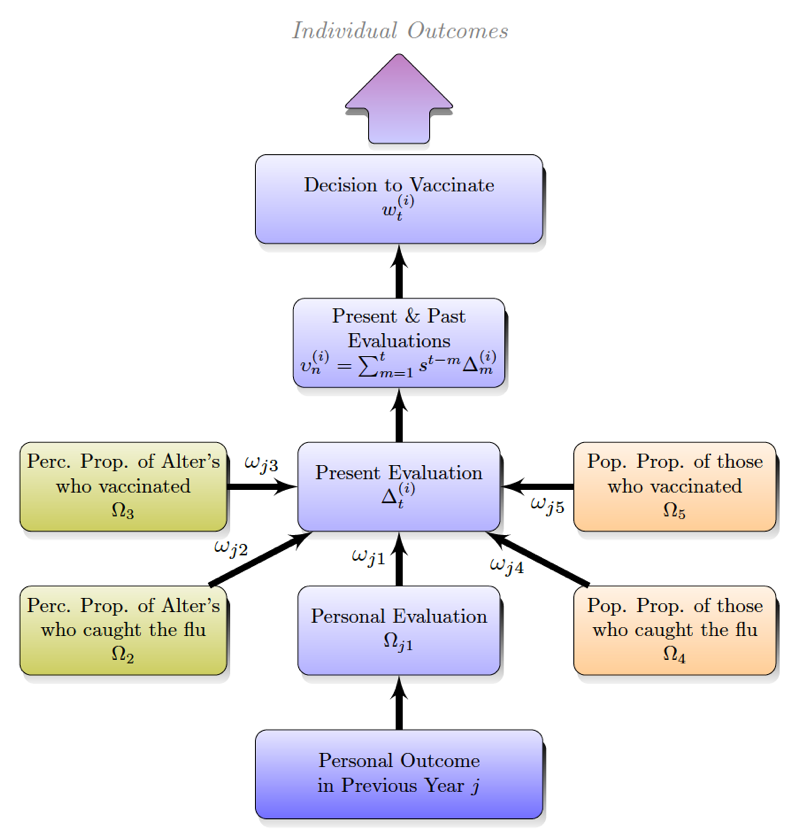}
		\caption{: Behavioral Model for Agents in Influenza ABM.}
		\label{fg:archBM}
\end{figure}

The  $ \beta $  coefficients are assumed to vary across socio-demographic population strata and measure the relative importance of each of these behavioral processes. Notice that  $ \beta _{c}$  is an array of regression coefficients. Individuals that are more likely to visit HCW before or during the flu season will likely be more influenced by their HCW and thus have a higher value of  $ \beta _{b}$. Instead, individuals belonging to households with low income may be more influenced by the cost of obtaining vaccination than by HCW’s recommendations and their past experiences, and thus have a higher cost related  $ \beta _{c}$.

In the model,  $ \phi _{n}^{ \left( i \right) }$  is determined by the influence of an HCW  $k$  in recommending his patient  $i$  to get vaccinated. HCWs follow guidelines in recommending vaccination. However, their effort in recommending vaccination is determined by their own perceptions that, in part, change with time. The model assumes that the perceptions of HCWs are formed from the population level cumulative incidence, the overall vaccination coverage, and the observation of the number of patients they visit every year with influenza.

The full ABM considers all three factors. However, for this project, we decided to use a simplified version of the behavioral model which considers a homogenous population that makes their vaccination decisions exclusively based on the present and past evaluations of their vaccination decisions and outcomes. As such in this simplified model we only consider the agent’s intrinsic adaptation and learning factor, and set the value  $ \beta _{b}$  to 1. Hence, agents are not influenced by recommendations made by their physicians. Moreover, the simplified model that we are considering here also removes other details that are important in the real system. For example, as mentioned previously the full model allows agents to be infected with a niILI and they could think that they caught the flu when instead they caught a niILI (or vice-versa). These details are not considered in the simplified model.

To model the agent’s intrinsic adaptation and learning, at the \textit{end} of season  $n-1$ , each individual evaluates the choice of either being vaccinated or not. Their evaluation is quantified by a variable  $ \Delta _{n-1}^{ \left( i \right) }$  ranging in  $ \left[ 0,1 \right] $ . Its value is large when the individual perceives the benefit of having been vaccinated. In the full ABM, this depends on three evaluation criteria:

\begin{enumerate}
	\item  $PE$ :\textit{ personal} and \textit{household} evaluations;
	\item  $SN$ \textbf{:} perceived \textit{local} decisions and outcomes in their social-network (SN); and
	\item  $MsM$ : perceived \textit{global} decisions and outcomes as reported by mainstream media (MsM).
\end{enumerate}

The importance placed on each of these levels of evaluation depends on three weights  $ \omega _{PE},~  \omega _{SN}~$, and  $ \omega _{MsM}$ ,\ which add to 1. In the simplified model,  we remove the MsM evaluation criteria. Moreover, the PE evaluation criterion is simplified to only depend on personal decisions and outcomes and not on household outcomes. Hence, the agent’s intrinsic adaptation and learning only depends on the PE and the SN evaluation criteria.

An agent’s PE generates a value  $ \Delta _{PE}^{ \left( i \right) }$  ranging in  $ \left[ 0,1 \right] $  which depend on four possible decision-outcome combinations, and result from (i) whether or not they were vaccinated, and (ii) whether or not they were infected by the end of the season. For example, if agent i vaccinated but still caught the flu, his/her value for  $ \Delta _{PE}^{ \left( i \right) }$ \ would\ be low and close to 0. This means that in absence of any information other than his/her most recent decision and outcome with influenza and its vaccine, his/her propensity to vaccinate again is low.   On the other hand if s/he did not vaccinate and caught the flu then  $ \Delta _{PE}^{ \left( i \right) }$  would be very high and usually set to equally 1. This means that in absence of any information, his/her propensity to vaccinate in the flowing season is very high. Similarly, an agent’s SN evaluation generates a value  $ \Delta _{SN}^{ \left( i \right) }$
ranging in $ \left[ 0,1 \right] $ \ which depends on the perceived proportion of alters in the social network that experienced each of the four possible decision-outcome combinations. For example, if a large majority of alters in agent $i$’s social network did not vaccinate and caught the flu, the value of   $ \Delta _{SN}^{ \left( i \right) }$  would be high and close to $1$.

In our simplified model, the value of  $ \Delta _{n-1}^{ \left( i \right) }$ is obtained by the weighted sum of  $ \Delta _{PE}^{ \left( i \right) }$ \ \ and   $ \Delta _{SN}^{ \left( i \right) }$ .

 $ \Delta _{n-1}^{ \left( i \right) }= \omega _{PE} \Delta _{PE}^{ \left( i \right) }+ \omega _{SN} \Delta _{SN}^{ \left( i \right) }$ .

Therefore, even if agent i did not vaccinate and did not catch the flu, leading to a low  $ \Delta _{PE}^{ \left( i \right) }$ \  value, if his/her  $ \Delta _{SN}^{ \left( i \right) }$  is high, s/he may decide to vaccinate in the following flu season. The interpretation here is that in absence of any information from other past influenza seasons, the agent will judge how lucky they were not to have caught the flu in the present season by his/her perception\ of how many alters in his/her social network did not vaccinate and caught the flu.  The actual values for  $ \Delta _{PE}^{ \left( i \right) }$  and  $ \Delta _{SN}^{ \left( i \right) }$ which result from the various decision outcomes of the agent and his/her alters can be set by tunable parameter values.

The value  $ \upsilon _{n}^{ \left( i \right) }$ describing an agent’s intrinsic adaptation and learning depends on both the most recent flu season and resulting value for  $ \Delta _{n-1}^{ \left( i \right) }$ , as well as past flu seasons resulting in values for  $ \Delta _{m}^{ \left( i \right) }$ for  $m<n-1$ . Each individual remembers and weights previous outcomes and their evaluations with respect to the present outcome. The weighted sum of past years’ evaluations defines individual  $i$ ’s pro-vaccination experience  $V_{n}^{ \left( i \right) }$  for the present season and is given by

 $V_{n}^{ \left( i \right) }=sV_{n-1}^{ \left( i \right) }+ \Delta _{n-1}^{ \left( i \right) }$ .

The parameter  $s$  ranges in  $ \left[ 0,1 \right] $  and discounts the previous year’s outcome with respect to the outcome of the present year. When  $s$  is equal to 0, individuals completely ignore the outcome of previous years. When  $s$  is equal to 1, individuals give equal weight to the outcomes from previous years as the present outcome. For example, when  $s$  is equal to 1, and\  if  $ \Delta _{n-1}^{ \left( i \right) }$  only took discrete values 0 or 1, the value  $V_{n}^{ \left( i \right) }$  would be an integer number representing a tally of the number of years that vaccination was or wound have benefitted agent i. When  $s<1$ ,\ and for continuous values of   $ \Delta _{n-1}^{ \left( i \right) }$  in  $ \left[ 0,1 \right] $ \ ,   $V_{n}^{ \left( i \right) }$  represents an exponential weighted moving average of the evaluations  $ \Delta _{n}^{ \left( i \right) }.~$ Finally, the probability that agent i will use to decide whether or not to vaccinate in the next flu season is found by normalizing  $V_{n}^{ \left( i \right) }$ using its maximum possible value. based only on the adaptive learning process is given by

 $$ \upsilon _{n}^{ \left( i \right) }=V_{n}^{ \left( i \right) }/N \left( s \right) , $$

where  $N \left( s \right) = \left( 1-s^{n} \right) / \left( 1-s \right) $ .

\subsection*{ABM Influenza Transmission Model}
\begin{figure}[htb]
	  \centering
		\includegraphics[width=0.35\textwidth,keepaspectratio]{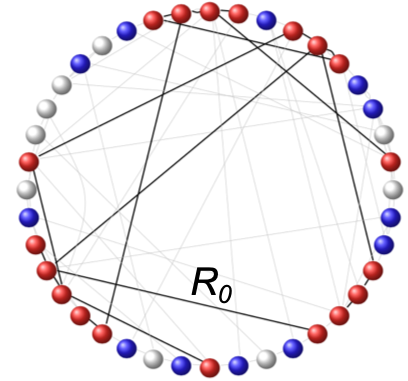}
		\caption{SIR percolation model on the Network.}
		\label{fg:sirperc}
\end{figure}

Influenza transmission is based on an SIR model over the social network structure that considers both node and edge percolation. Individuals, represented as nodes on the network, begin in either one of the three states: susceptible, infected, or recovered. At the start of the season, some individuals decide to seek vaccination. This is determined according to the probabilities  $w_{i}^{ \left( i \right) }$ ’s described previously. Based on the vaccine efficacy, we deactivate edges from the network that connect to nodes representing immune individuals who chose to get vaccinated. The degree of protection conferred by an influenza vaccine is assumed to be 50$\%$  to 70$\%$  and can vary from season to season. The lower range in efficacy will be applied to the elderly in our population. Figure~\ref{fg:sirperc} below provides an illustration where blue nodes are immune due to vaccination.

During an influenza epidemic, not all individuals in the network that remain susceptible to infection will become infected. Depending on the influenza transmissibility, some individuals will come in contact with infectious individuals but still avoid infection. Therefore, in the model influenza transmission is allowed to propagate only through the active edges of the network (shown in bold). An edge that connects two susceptible individuals  $i$  and  $j$  stays active with probability  $T_{ij}= \left[ 1-\exp  \left( - \lambda _{ij}/ \gamma  \right)  \right] $  which depends on

\begin{enumerate}
	\item influenza transmissibility rate ( $ \beta _{ij}$ ), and
	\item the duration for which either individual is infectious and continues interacting socially in the network ($\gamma^{-1}$).
\end{enumerate}

The infectiousness duration depends on the probability that either individual is symptomatic and stays home in the event of being infected (i.e., social-distancing). Influenza transmissibility, infectiousness duration, probability of being asymptomatic, and social-distancing depends on socio-demographic attributes of  $i$  and  $j$  have been parameterized from the literature. The basic reproductive number resulting from this model is  $R_{0}= \langle \langle k \rangle _{m} \rangle \langle T_{ij} \rangle $  where  $\langle T_{ij} \rangle $  is the average transmission probability in our resulting network.

\section*{Appendix B: Statement of Modified MAC RL Algorithm}
Here we outline the learning algorithm used in the ABM experiments. We take inspiration from \cite{lowe2017multi} and implement an adapted RL algorithm that aims to address the complexities specific to multi-agent learning, the Multi-agent Actor-Critic (MAC) algorithm (see below). It is based on the same principle underlying the Multi-Agent Deep Deterministic Policy Gradient (MADDPG) algorithm benchmarked in \cite{lowe2017multi}. It aims to enable limited time information-sharing via a globally-accessible full-observation oracle or central Q-critic during the training phase. The central Q-critic is more informative about the value of action-state pairs. The use of this augmented oracle leads to more stable learning.

Algorithm~\ref{alg:macvar} also distinguishes between information-sharing during the test vs. training phases. The individual agents’ access to central Q-critic is limited to just the training phase. For continued adaptation after model deployment, we can still update the agent’s model using value estimates from its local critic instead of the central Q-critic. We will use the following notation as a compact way to denote a population variable: $\vec{d}_t = \{d_{i,t}\}_{i=1}^N$.

\begin{algorithm}[ht]
   \caption{Variant on Multiagent Actor-Critic (MAC) Algorithm}
   \label{alg:macvar}
\begin{algorithmic}
   \STATE {\bfseries Input:} $\{s_{i,t}, a_{i,t}, r_{i,t}, s_{i,t+1} \}_{t=0}^T$ episode experience tuple for $N$ agents
   \REPEAT
   		\STATE Initialize $s_{i,0} \quad \forall i \in \{1,\cdots, N\}$.
	   \FOR{each $i \in \{1,\cdots, N\}$}
	   	\FOR{each $t \in \{0,\cdots, T-1\}$}
	   		\STATE Compute $i^{th}$ advantage:
	   		\STATE $Adv_{i,t} = r_{i,t} + \gamma Q_i(\vec{a}_{t+1}, \vec{s}_{t+1}; \Phi_{t}) - Q_i(\vec{a}_{t}, \vec{s}_{t}; \Phi_{t}) $
	   		\STATE Apply $i^{th}$ actor's policy gradient ascent update:
	   		\STATE $\Theta_{t+1} = \Theta_{t} + \alpha_\pi \nabla_{\Theta} \log \pi_i(a|s_t, \Theta_t)$
	   		\IF{Training}
	   			\STATE Apply central-Q gradient descent update:
	   			\STATE $\Phi_{t+1} = \Phi_{t}  - \alpha_Q \nabla_\Phi Adv_{i,t}^2$
		   	\ENDIF
	   		\IF{Deployed}
	   			\STATE Apply $i^{th}$ critic's gradient ascent update using local advantage:
	   			\STATE $Adv(s_{i,t}) = r_{i,t} + \gamma v(s_{i,t+1},\omega_t) - v(s_{i,t},\omega_t)$
	   			\STATE $\omega_{t+1} = \omega_{t}  + \alpha_v Adv(s_{i,t}) \nabla_\omega v(s_{i,t},\omega_t)$
		   	\ENDIF
	   	\ENDFOR
	   \ENDFOR
   \UNTIL{simulation ends}
\end{algorithmic}
\end{algorithm}





\end{document}